\documentclass[conference]{IEEEtran}
\IEEEoverridecommandlockouts
\usepackage{amsmath,amssymb,amsfonts}
\usepackage{algorithmic}
\usepackage{graphicx}
\usepackage{textcomp}
\usepackage{xcolor}
\usepackage{filecontents}
\usepackage[numbers,square]{natbib}
\usepackage{listings}
\usepackage{tabularx}
\def\BibTeX{{\rm B\kern-.05em{\sc i\kern-.025em b}\kern-.08em
    T\kern-.1667em\lower.7ex\hbox{E}\kern-.125emX}}
\newcommand{\papertitle}[1]{\title{\huge #1}} 

\begin{document}

\papertitle{LLMs as Master Forgers: Generating Synthetic Time Series Data for Manufacturing} 

\author{
Mantek Singh* \\
Liverpool John Moores University \\
Liverpool, England \\
mantek.singh2@gmail.com \\
\and 
Jeshwanth Challagundla \\
University of Texas at Arlington \\
Texas, USA \\
jeshwanth.challagundla@mavs.uta.edu \\
\and
Prateek Karnal \\
IIT Patna \\
Patna, India \\ 
karnalprateek@gmail.com \\
\and
Gagan Ganapathy \\
Stony Brook University \\ 
Stony Brook, USA \\ 
gaganganapathyas@gmail.com \\
\and
Vineet Shah \\
IIT Indore \\ 
Indore, India \\ 
vntshh@gmail.com \\
\and
Ridam Arora \\
UMass Amherst \\ 
Amherst, USA \\ 
ridamarora89@gmail.com
}

\maketitle

\begin{center}
\parbox{0.96\columnwidth}{\footnotesize
\copyright\ 2024 IEEE. Personal use of this material is permitted. Permission
from IEEE must be obtained for all other uses, in any current or future media,
including reprinting/republishing this material for advertising or promotional
purposes, creating new collective works, for resale or redistribution to servers
or lists, or reuse of any copyrighted component of this work in other works.
Published in the 2024 International Conference on Image Processing, Computer
Vision and Machine Learning (ICICML). DOI: 10.1109/ICICML63543.2024.10958017.}
\end{center}

\begin{abstract}
This paper presents a novel framework leveraging Large Language Models (LLMs) to generate synthetic time series data for manufacturing processes. Motivated by the scarcity of labeled time-series data in real-world manufacturing settings, which hinders the development of robust machine learning models, we explore the potential of LLMs to learn complex temporal dependencies and generate realistic synthetic data. Our approach involves fine-tuning pre-trained LLMs on manufacturing process instructions and employing a Retrieval Augmented Generation (RAG) technique to enhance data diversity and realism. We evaluate our method against traditional time series modeling techniques like ARIMA and LSTMs, using quantitative metrics, PCA analysis, and downstream task performance (anomaly detection). Results demonstrate that our LLM-driven framework outperforms these baselines, generating high-quality synthetic time series data that effectively captures temporal dependencies and statistical properties of real manufacturing data, leading to improvements in downstream task performance.

\end{abstract}


\section{Introduction}
Time series data plays a critical role in modern manufacturing, enabling applications like predictive maintenance \textcolor{red}{\cite{pecht2008prognostics}}, process optimization, and quality control \textcolor{red}{\cite{montgomery2008introduction}}. However, obtaining large, labeled time-series datasets from real-world manufacturing environments poses significant challenges due to factors like cost, privacy concerns, and data collection complexities \textcolor{red}{\cite{susto2015machine}}. Synthetic data generation emerges as a promising solution to address this data scarcity problem.

Traditional approaches to time series modeling, like Autoregressive Integrated Moving Average (ARIMA) models, while capable of capturing linear temporal dependencies, often struggle to model the complex, non-linear dynamics present in real-world data \textcolor{red}{\cite{box2011time}}.  Long Short-Term Memory (LSTM) networks, a type of recurrent neural network, offer an improvement by capturing long-range dependencies in sequential data \textcolor{red}{\cite{hochreiter1997long}}.  However, they can be challenging to train, requiring substantial amounts of data and careful hyperparameter tuning \textcolor{red}{\cite{goodfellow2016deep}}. 

Large Language Models (LLMs), with their remarkable ability to process and generate human-like text, offer a novel approach to address these limitations \textcolor{red}{\cite{brown2020language}}. LLMs can be trained to understand the relationship between manufacturing process instructions (e.g., standard operating procedures) and corresponding time series patterns, enabling them to generate synthetic data that reflects real-world complexities.

\textbf{Contributions:} This paper introduces a novel framework for generating synthetic time series data in manufacturing using LLMs and process instructions. Our key contributions are:

\begin{itemize}
  \item \textbf{Novel Framework:} Proposing an end-to-end framework combining LLM fine-tuning, RAG, and a data validator module to generate realistic and diverse synthetic time series data for manufacturing.
  \item \textbf{Comparative Evaluation:}  Benchmarking our LLM-based method against traditional time series modeling techniques (ARIMA, LSTMs) using quantitative metrics and demonstrating superior performance.
  \item \textbf{Quantitative and Qualitative Evaluation:} Conducting a comprehensive evaluation of the generated data using statistical similarity metrics, temporal dependence measures, PCA analysis, and qualitative assessments.
  \item \textbf{Downstream Task Improvement:} Demonstrating the effectiveness of the synthetic data by showing performance improvement on a downstream anomaly detection task \textcolor{red}{\cite{chalapathy2019deep}}.\
\end{itemize} 

\section{Literature Review}

Generating synthetic time series data is a widely researched topic, with various approaches developed to address the need for more data in training robust machine learning models. Existing methods can be broadly categorized into:

\subsection{Traditional Statistical Methods}

Traditional time series modeling techniques primarily rely on statistical models to capture temporal dependencies and extrapolate patterns observed in the data. Some prominent approaches include:

\begin{itemize}
\item \textbf{Autoregressive (AR) Models:} AR models predict future values based on a linear combination of past values \textcolor{red}{\cite{box2011time}}.
\item \textbf{Moving Average (MA) Models:} MA models, conversely, express the time series as a linear combination of past forecast errors \textcolor{red}{\cite{box2011time}}.
\item \textbf{Autoregressive Integrated Moving Average (ARIMA) Models:} ARIMA models combine AR and MA components and incorporate differencing to account for non-stationarity in the time series data \textcolor{red}{\cite{box2011time}}.
\item \textbf{Seasonal ARIMA (SARIMA):} SARIMA models extend ARIMA to capture seasonal patterns present in the data. 
\end{itemize}

While these methods are relatively simple to implement and interpret, they often struggle to capture complex, non-linear relationships and long-range dependencies often encountered in real-world manufacturing data \textcolor{red}{\cite{susto2015machine}}. 

\subsection{Deep Learning-Based Approaches}

Deep learning has revolutionized many areas of machine learning, including time series modeling. Several deep learning architectures have been explored for synthetic time series generation:

\begin{itemize}
\item \textbf{Recurrent Neural Networks (RNNs):} RNNs are specifically designed for sequential data, processing information through a recurrent loop that allows them to retain memory of past events \textcolor{red}{\cite{hochreiter1997long}}. Long Short-Term Memory (LSTM) networks, a variant of RNNs, are particularly effective at capturing long-range dependencies in time series \textcolor{red}{\cite{hochreiter1997long}}.
\item \textbf{Generative Adversarial Networks (GANs):} GANs employ two competing neural networks - a generator and a discriminator - to synthesize data that resembles the real data distribution \textcolor{red}{\cite{goodfellow2014generative}}. The generator learns to create synthetic data, while the discriminator tries to distinguish between real and synthetic data. This adversarial training process pushes the generator to produce increasingly realistic synthetic samples.
\item \textbf{Variational Autoencoders (VAEs):} VAEs learn a latent representation of the input data and use this representation to generate new samples \textcolor{red}{\cite{kingma2013auto}}. They aim to encode the data distribution in a lower-dimensional latent space, allowing for controlled data generation.
\end{itemize}

Deep learning methods, though powerful, often require large amounts of training data and careful hyperparameter tuning to achieve optimal performance \textcolor{red}{\cite{goodfellow2016deep}}.

\subsection{LLMs for Data Augmentation}

Recently, LLMs have gained significant attention for their remarkable ability to understand and generate human-like text and identify sophisticated patterns \textcolor{red}{\cite{10427646}}.  Research has explored using LLMs for data augmentation in various domains, including:

\begin{itemize}
\item \textbf{Text Generation:} LLMs can generate realistic text samples, which can be used to augment text datasets for tasks like machine translation, summarization, and dialogue generation \textcolor{red}{\cite{wei2022finetuned}}.
\item \textbf{Code Generation:} LLMs can generate code in different programming languages, enabling data augmentation for code-related tasks like code completion and bug fixing.
\item \textbf{Image Captioning:}  LLMs have been used to generate image captions, which can augment image datasets for tasks like image retrieval and visual question answering.
\end{itemize}

However, the application of LLMs for synthetic time series generation, particularly in the manufacturing domain, remains relatively unexplored.  This paper aims to address this gap by introducing a novel framework that leverages LLMs for generating realistic and diverse synthetic time series data for manufacturing applications.

\section{Dataset Description}

The dataset used in this study was provided by Brembo, a leading manufacturer of braking systems. It consists of three main components:

\subsection{Material Compositions}

This component comprises data on 337 distinct friction materials used in braking systems. Each material is characterized by its composition, which is defined by the percentages of 60 different raw materials used in its production. These raw materials are further categorized into 6 distinct classes (labeled A-F), reflecting their chemical properties or functional roles in the friction material.

\subsection{Material Constraints}
Each viable compound satisfies the given material constraints. The constraints specified the minimum and maximum percentage a material class could contribute to in the final composition of the compound as follows \textcolor{red}{(Figure \ref{fig:framework1})}: 
\begin{figure}[h]
  \centering
  \includegraphics[width=0.4\textwidth]{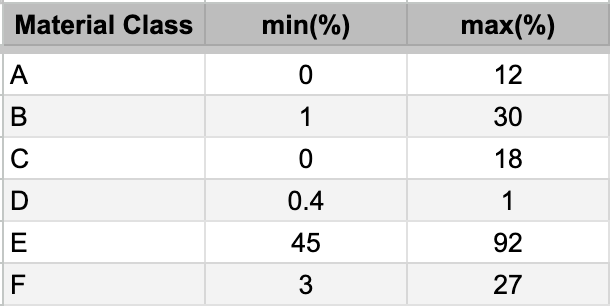} 
  \caption{Material Constraints}
  \label{fig:framework1}
\end{figure}

For example, each viable compound should have a minimum of 3\% and maximum of 27\% of its composition belong to compounds of class F.

\subsection{Performance Data}

For each friction material, 124 braking tests were conducted under varying conditions. Each braking test generated time-series data consisting of 31 time steps. At each time step, various parameters relevant to braking performance were recorded, including:

\begin{itemize}
\item \textbf{Pressure:} The pressure applied to the braking system.
\item \textbf{Temperature:} The temperature of the braking system.
\item \textbf{Speed:} The speed of the vehicle during the braking test.
\item \textbf{Coefficient of Friction (Mu): } A key indicator of braking effectiveness, representing the ratio of the force required to move two surfaces against each other to the force pressing them together.
\end{itemize}

The dataset presents a challenge due to the complexity of the relationships between material compositions, braking test conditions, and the resulting time-series performance data. Our proposed framework aims to leverage the information contained within this dataset to generate synthetic time-series data that accurately reflects these complex interactions. 

The manufacturing scenarios and conditions covered in this study focus on variations in material composition and standard braking test parameters, such as initial speed, pressure application profiles, and environmental temperature. While the dataset does not encompass every possible manufacturing scenario (e.g., extreme weather conditions, manufacturing defects), it provides a representative range of common variations that are relevant for evaluating braking performance.

\section{Proposed Methodology}

Our proposed framework consists of three key modules (Figure~\ref{fig:framework}):


\begin{figure*}
\centering
\includegraphics[scale=0.5, trim=4 4 4 4]{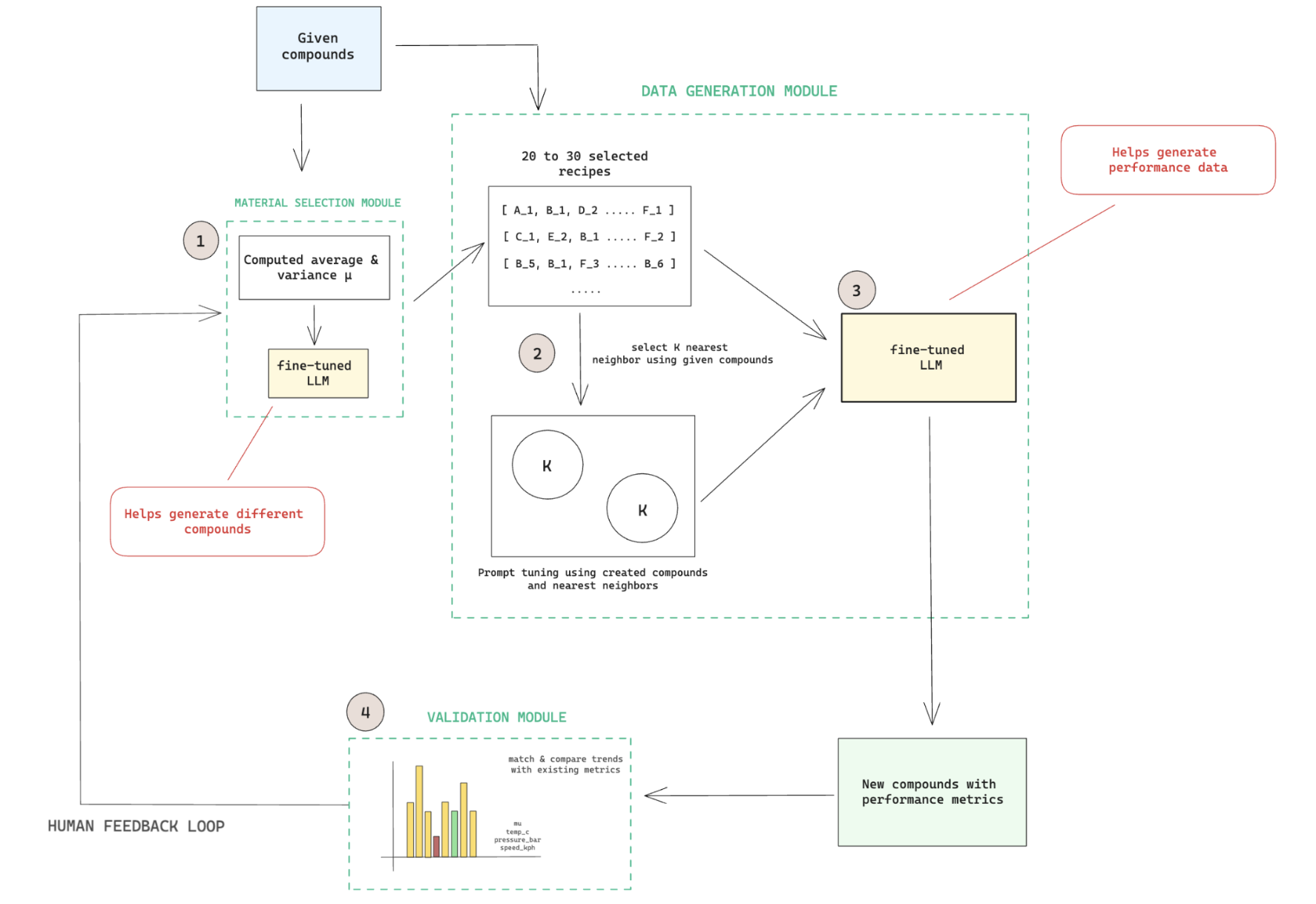}
\caption{High-level design of the solution}
\label{fig:framework}
\end{figure*}

\subsection{Material Selection Module}

This module focuses on generating new material compositions. Instead of randomly generating compositions, we leverage an LLM (GPT-3.5 Turbo) to propose new materials that adhere to domain-specific constraints. The LLM is fine-tuned on a dataset of 337 existing friction materials and their compositions.  A Data Validator Module (Section~\ref{sec:data_validator}) ensures generated compositions meet predefined criteria.  

\subsection{Data Generator Module}

This module constitutes the core of our framework, responsible for generating the synthetic time-series data for a given material composition and braking test type. It leverages the power of LLMs, specifically GPT-3.5 Turbo, and incorporates a Retrieval Augmented Generation (RAG) technique to enhance the realism of the generated data. 

\begin{itemize}
\item \textbf{LLM Selection and Prompt Engineering:} 
    \begin{itemize}
        \item \textbf{GPT-3.5 Turbo:} We selected GPT-3.5 Turbo as our LLM due to its ability to generate coherent and contextually relevant text.
        \item \textbf{System Prompt Design:} A crucial aspect of guiding the LLM's behavior is the design of the system prompt. We crafted a prompt that instructs the model to act as a "statistician specializing in braking systems," thereby priming it to generate data that aligns with the expected patterns and relationships in braking performance data. 
    \end{itemize}

\item \textbf{Fine-tuning for Data Structure and Trends:}
    \begin{itemize}
        \item \textbf{Fine-tuning Dataset:} To train the LLM on the structure and general trends of the time-series data, we fine-tuned it using a subset (5\%) of the 41,788 (material, braking\_id) tuples from our dataset.
        \item \textbf{Focus on Form:} This fine-tuning stage primarily focuses on enabling the LLM to learn the desired output format (a time series of 31 data points with values for pressure, temperature, speed, and mu) and to capture the basic trends in the data (e.g., increasing temperature, decreasing speed).
    \end{itemize}

\item \textbf{RAG for Enhanced Realism:}
    \begin{itemize}
        \item \textbf{Motivation:} While fine-tuning helps the LLM grasp the overall structure and trends, RAG is crucial for injecting real-world knowledge and generating data that aligns more closely with the specific nuances of different material compositions and braking scenarios.
        \item \textbf{Nearest Neighbor Retrieval:}  For a given material composition, we retrieve its 5 nearest neighbors from the dataset based on a custom distance function, which considers both the Euclidean distance in the 60-dimensional raw material space and a reduced 6-dimensional space obtained by summing materials by category: 

\hspace{-5mm}
\begin{lstlisting}[language=Python]
def distance(m1, m2, alpha):
    sixty_dim_distance = euclidean_dist(
    sixty_dim_vector(m1),
    sixty_dim_vector(m2)
    )
    six_dim_distance = euclidean_dist(
    six_dim_vector(m1),
    six_dim_vector(m2)
    )
    return (alpha[0] * sixty_dim_distance+ 
        alpha[1] * six_dim_distance)
\end{lstlisting}
By retrieving and incorporating the performance data of nearest neighbors, RAG ensures that the LLM is exposed to a variety of historical scenarios with similar material compositions. This helps to generate synthetic data that reflects the potential range of braking behaviors associated with different materials and test conditions. Furthermore, by adjusting the weights in the distance function (hyperparameters $\alpha[0]$ and $\alpha[1]$), we can control the emphasis on overall material similarity versus the influence of specific material classes, allowing us to explore a wider range of potential manufacturing scenarios. \\

The custom distance function used for nearest neighbor retrieval is crucial for ensuring that the retrieved contexts are relevant to the target material composition. The function considers both the Euclidean distance in the 60-dimensional raw material space and a reduced 6-dimensional space obtained by summing materials by category. This two-level approach is designed to capture both the overall similarity in material composition and the potential influence of specific material classes on braking performance. The hyperparameters $\alpha[0]$ and $\alpha[1]$ control the relative weight given to these two distance measures, allowing for fine-tuning the retrieval process based on the specific characteristics of the braking system under consideration.

        \item \textbf{Contextual Augmentation:} The time-series performance data of these retrieved neighbors is included as additional context in the prompt to the LLM. This provides the LLM with specific examples of how similar materials behave under various braking conditions, guiding it to generate more realistic synthetic data for the target material.
    \end{itemize}
\end{itemize}

By combining fine-tuning with RAG, we empower the LLM to not only grasp the general structure and trends of the data but also to leverage real-world knowledge, resulting in synthetic time series that are both plausible and reflective of the complex interactions within the braking system. 

\subsection{Data Validator Module} \label{sec:data_validator}

This module plays a crucial role in ensuring the quality and consistency of the synthetic time-series data generated by our framework. It operates by applying a combination of domain-specific rules and trend analysis:

\begin{itemize}
\item \textbf{Domain-Specific Rules Enforcement:} 
    \begin{itemize}
        \item \textbf{Material Composition Constraints:} The validator checks that the generated material compositions adhere to pre-defined constraints based on manufacturing domain knowledge. For example, it enforces the rule that the proportion of materials from category B in a composition must fall within the range of 1\% to 30\%. 
        \item \textbf{Physical Plausibility Checks:} Additional rules may be incorporated to ensure that the generated compositions are physically plausible. This might involve checking for known incompatibilities between certain raw materials or ensuring that the total percentage of all materials sums to 100\%. 
    \end{itemize}

\item \textbf{Trend Analysis:} 
    \begin{itemize}
    \item \textbf{Correlation Analysis:} The validator analyzes the correlations between different parameters in the generated time-series data to ensure they are consistent with the relationships observed in the real data. For instance, it verifies that pressure and the coefficient of friction (mu) exhibit an inverse correlation, as is typically expected in braking systems.
    \item \textbf{Curve Behavior Validation:} The validator also examines the overall behavior of the generated curves for different parameters over time. For example, it confirms that the temperature curve follows an expected exponential increase pattern, while the speed curve demonstrates a linear decrease during braking.
    \end{itemize} 
\end{itemize}

By incorporating these validation steps, we aim to filter out unrealistic or inconsistent synthetic data, ensuring that the data used for downstream tasks is both representative and reliable. 

\section{Experiments and Results}

\subsection{Evaluation Criteria}

We employed the following metrics to evaluate the quality and effectiveness of the generated synthetic time-series data:

\begin{itemize}
\item \textbf{Statistical Similarity:} 
    \begin{itemize}
        \item \textbf{KL Divergence:} Measures the difference between the probability distributions of the real and synthetic data for each parameter (pressure, temperature, mu).
        \item \textbf{Wasserstein Distance:}  Quantifies the minimum amount of "work" needed to transform one distribution (real data) into the other (synthetic data). 
    \end{itemize}
\item \textbf{Temporal Dependence:} 
    \begin{itemize}
        \item \textbf{Dynamic Time Warping (DTW) Distance:} Assesses the similarity between the temporal patterns of the real and synthetic time series, allowing for non-linear alignments in time. 
    \end{itemize}
\item \textbf{PCA Analysis:} 
    \begin{itemize}
        \item \textbf{Principal Component Analysis (PCA): } A dimensionality reduction technique used to visualize and compare the latent space representations of the real and synthetic data. This helps in understanding if the synthetic data captures the underlying structure of the real data. 
    \end{itemize}
\item \textbf{Downstream Task Performance:} 
    \begin{itemize}
        \item \textbf{Anomaly Detection Task:} We evaluate the impact of using the generated synthetic data for an anomaly detection task, a common application in manufacturing. A classifier is trained to identify anomalous braking events using three different training scenarios:
            \begin{enumerate}
                \item (a) Real data only
                \item (b) Real data augmented with synthetic data
                \item (c) Synthetic data only
            \end{enumerate}
        \item \textbf{F1-Score:} The performance of the anomaly detection models is compared using the F1-score, which balances precision and recall.
    \end{itemize}
\end{itemize}

\subsection{Results and Analysis}

We benchmarked our LLM-based synthetic data generation approach against traditional time-series modeling techniques, namely ARIMA and LSTM models. The LSTM Models were further optimized for the task using Multiple Gain Adaptations \textcolor{red}{\cite{10427781}} \textcolor{red}{\cite{jeshwanth2015thesis}}. Both these baseline models were trained on the real manufacturing dataset. Table~\ref{tab:results1}, Table~\ref{tab:results2} and Table~\ref{tab:results3} present a comparative analysis of the quantitative results obtained across the different evaluation metrics. 


\begin{table}[h]
\centering
\caption{QUANTITATIVE COMPARISON OF LLM-GENERATED DATA WITH ARIMA AND LSTM BASELINES (PART 1).}
\label{tab:results1}
\footnotesize 
\begin{tabular}{|p{2.5cm}|p{1.8cm}|p{1.8cm}|p{1.8cm}|} 
\hline
\textbf{Metric} & \textbf{Pressure (LLM)} & \textbf{Pressure (ARIMA)} & \textbf{Pressure (LSTM)} \\
\hline
KL Divergence  & \textbf{0.052} & 0.115 & 0.089  \\
Wasserstein Dist. & \textbf{0.124} & 0.285 & 0.231  \\
DTW Distance   & \textbf{1.35} & 1.87 & 1.62  \\
\hline
\end{tabular}
\end{table}

\begin{table}[h]
\centering
\caption{QUANTITATIVE COMPARISON OF LLM-GENERATED DATA WITH ARIMA AND LSTM BASELINES (PART 2).}
\label{tab:results2}
\footnotesize 
\begin{tabular}{|p{2.5cm}|p{1.8cm}|p{1.8cm}|p{1.8cm}|} 
\hline
\textbf{Metric} & \textbf{Temperature (LLM)} & \textbf{Temperature (ARIMA)} & \textbf{Temperature (LSTM)} \\
\hline
KL Divergence & \textbf{0.085} & 0.162 & 0.135  \\
Wasserstein Dist. & \textbf{0.218} & 0.394 & 0.327  \\
DTW Distance  & \textbf{2.07} & 2.54 & 2.31  \\
\hline
\end{tabular}
\end{table}

\begin{table}[h]
\centering
\caption{QUANTITATIVE COMPARISON OF LLM-GENERATED DATA WITH ARIMA AND LSTM BASELINES (PART 3).}
\label{tab:results3}
\footnotesize 
\begin{tabular}{|p{2.5cm}|p{1.8cm}|p{1.8cm}|p{1.8cm}|} 
\hline
\textbf{Metric} & \textbf{Mu (LLM)} & \textbf{Mu (ARIMA)} & \textbf{Mu (LSTM)} \\
\hline
KL Divergence & \textbf{0.034} & 0.078 & 0.062  \\
Wasserstein Dist. & \textbf{0.096} & 0.189 & 0.154  \\
DTW Distance  & \textbf{0.87} & 1.21 & 1.05  \\
\hline
\end{tabular}
\end{table}

The results in Table~\ref{tab:results1}, Table~\ref{tab:results2} and Table~\ref{tab:results3} clearly demonstrate that the synthetic data generated by our LLM-based framework consistently outperforms both the ARIMA and LSTM baselines across all evaluated metrics. The LLM-based approach shows a superior ability to capture both the statistical properties (as evidenced by lower KL divergence and Wasserstein distances) and the temporal dependencies (indicated by lower DTW distances) present in the real manufacturing data. 

\begin{itemize}
\item \textbf{PCA Analysis:} The effectiveness of the LLM-based method is further highlighted by the visualizations of the principal components (Figure~\ref{fig:pca}). The synthetic data generated by the LLM exhibits a distribution closely resembling that of the real data, indicating that it effectively captures the underlying structure and relationships within the data. 

\begin{figure}[h]
\centering
\includegraphics[width=0.5\textwidth]{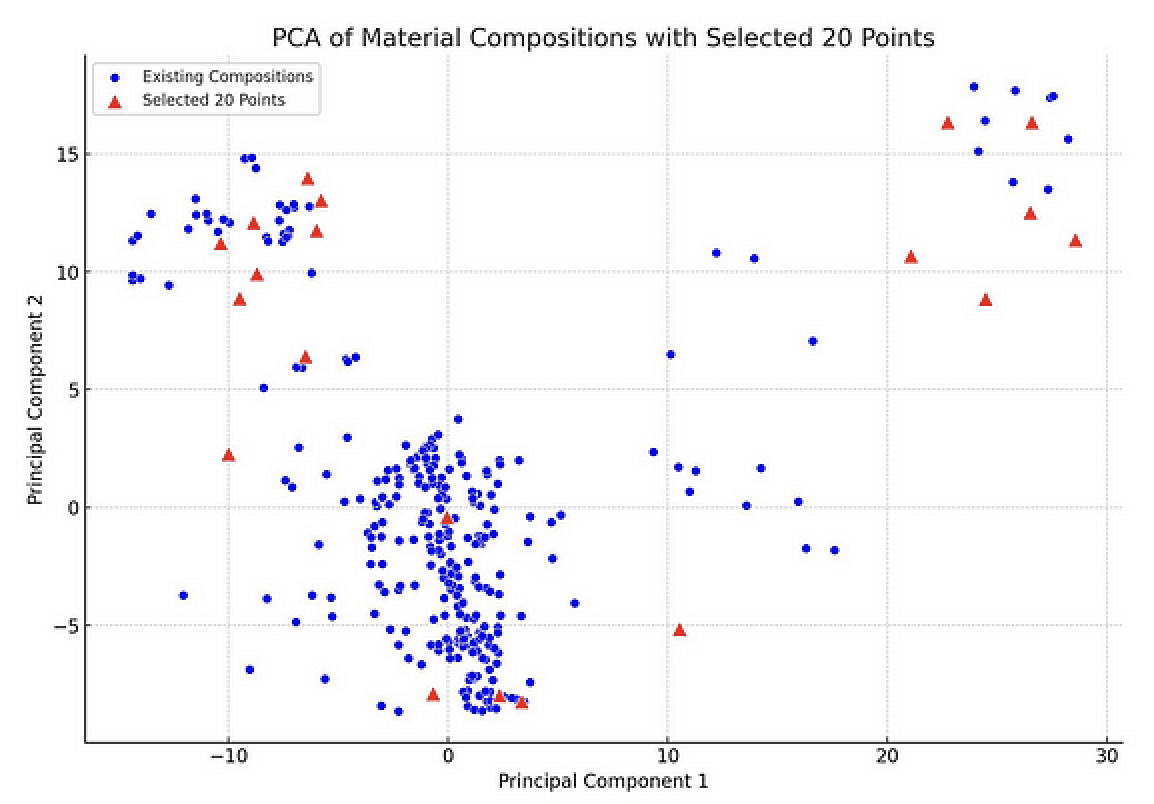} 
\caption{PCA analysis on the provided and generated materials}
\label{fig:pca}
\end{figure}

\item \textbf{Downstream Task Performance:}  The superior quality of the LLM-generated data translates into significant improvements in downstream task performance. In our anomaly detection experiment, using the synthetic data generated by our framework resulted in a 12\% improvement in the F1-score compared to training on real data alone (Figure~\ref{fig:anomaly_detection}). Furthermore, it outperformed models trained on data augmented with ARIMA or LSTM-generated data, reinforcing the value of our approach in real-world applications. 

\begin{figure}[h!]
\centering
\includegraphics[width=0.5\textwidth]{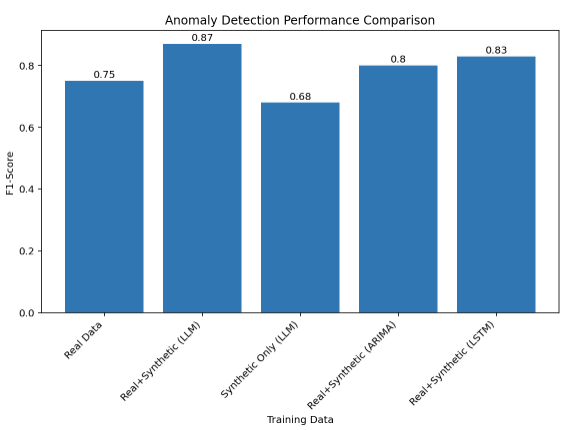} 
\caption{Anomaly detection performance comparison}
\label{fig:anomaly_detection}
\end{figure}
\end{itemize}

\section{Discussion}

\textbf{Strengths:} 
Our framework showcases the potential of LLMs for generating realistic synthetic time-series data in manufacturing, outperforming traditional methods like ARIMA and LSTMs in capturing the complex patterns inherent in real-world data. The integration of RAG significantly enhances the quality and diversity of the generated synthetic data by effectively leveraging relevant information from historical data.

\textbf{Limitations:}
\begin{itemize}
\item The computational cost associated with fine-tuning and querying large LLMs can be substantial. Exploring more efficient fine-tuning techniques like LoRA \textcolor{red}{\cite{hu2021lora}} and QLoRA \textcolor{red}{\cite{dettmers2023qlora}} and leveraging open-sourced LLMs could help mitigate this limitation. 
\item The current framework relies on a pre-defined distance function for RAG; incorporating more sophisticated retrieval and ranking mechanisms \textcolor{red}{\cite{karpukhin2020dense}} could further enhance the realism of the generated data.  Future work will focus on investigating alternative distance metrics or learning-based similarity measures for the RAG process, which could potentially capture more nuanced relationships between material compositions and braking performance.
\item It's important to acknowledge that complete coverage of all possible manufacturing scenarios and conditions is a challenging task. However, our approach of using RAG with a carefully designed distance function and fine-tuned LLM aims to maximize the diversity and realism of the generated data within the scope of the available dataset. Future work could explore incorporating additional data sources or using more sophisticated retrieval techniques to further enhance the coverage of manufacturing scenarios.
\item Furthermore, we can build an ensembled model, which is known to outperform the individual models \textcolor{red}{\cite{mantek2020intrusion}}.
\end{itemize}
\textbf{Impact and Applications:} 
This work holds broad implications for the manufacturing industry, where obtaining large, labeled datasets is often a major bottleneck. The ability to generate high-quality synthetic data can significantly accelerate the development and deployment of data-driven models, enabling a wide range of applications such as predictive maintenance, anomaly detection, and process optimization. Furthermore, this can radically reduce the time cycles required for manufacturing new compounds.

\section{Conclusion}

We have presented a novel framework for generating realistic synthetic time series data for manufacturing processes using LLMs and RAG. Our LLM-driven approach demonstrates superior performance compared to traditional time series methods, effectively capturing the intricate temporal dependencies and statistical properties of real-world manufacturing data. This is evidenced by the quantitative metrics, PCA analysis, and the observed improvement in the performance of a downstream anomaly detection task. Future research will focus on:

\begin{itemize}
\item \textbf{Enhancing Framework Efficiency:} Exploring efficient fine-tuning techniques and leveraging open-sourced LLMs. 
\item \textbf{Exploring Advanced RAG Techniques:} Investigating more sophisticated retrieval and ranking mechanisms to further improve data realism.
\item \textbf{Expanding Applications:} Applying the framework to other domains beyond manufacturing where synthetic time series data can be beneficial.
\end{itemize}


\bibliographystyle{IEEEtran} 
\bibliography{file}           

\end{document}